\documentclass[journal]{IEEEtran}
\usepackage{amsfonts}
\usepackage{amsmath}
\usepackage{graphicx}
\usepackage{cleveref}
\usepackage{color,soul}
\usepackage[export]{adjustbox}
\usepackage{cite}
\usepackage{multirow}
\begin{document}

\title{Deep Active Learning for Multi-Label Classification of Remote Sensing Images}

\author{Lars~Möllenbrok~\IEEEmembership{~Member,~IEEE}, Gencer Sumbul,~\IEEEmembership{Graduate Student~Member,~IEEE}, Begüm~Demir,~\IEEEmembership{Senior~Member,~IEEE}
\thanks{L. Möllenbrok, G. Sumbul and B. Demir are with the Faculty of Electrical Engineering and Computer Science, Technische Universit{\"a}t Berlin, 10623 Berlin,
Germany, also with the BIFOLD - Berlin Institute for the Foundations of Learning and Data, 10623 Berlin, Germany. Email: \mbox{lars.moellenbrok@tu-berlin.de}, \mbox{gencer.suembuel@tu-berlin.de}, \mbox{demir@tu-berlin.de}. This work is supported by the European Research Council (ERC) through the ERC-2017-STG BigEarth Project under Grant 759764 and by the German Ministry for Education and Research as BIFOLD - Berlin Institute for the Foundations of Learning and Data (01IS18025A).}
}

\markboth{Journal of \LaTeX\ Class Files,~Vol.~13, No.~9, September~2014}%
{Shell \MakeLowercase{\textit{et al.}}: Bare Demo of IEEEtran.cls for Journals}

\maketitle

\begin{abstract}
In this letter, we introduce deep active learning (AL) for multi-label image classification (MLC) problems in remote sensing (RS). In particular, we investigate the effectiveness of several AL query functions for MLC of RS images. Unlike the existing AL query functions (which are defined for single-label classification or semantic segmentation problems), each query function in this paper is based on the evaluation of two criteria: i) multi-label uncertainty; and ii) multi-label diversity. The multi-label uncertainty criterion is associated to the confidence of the deep neural networks (DNNs) in correctly assigning multi-labels to each image. To assess this criterion, we investigate three strategies: i) learning multi-label loss ordering; ii) measuring temporal discrepancy of multi-label predictions; and iii) measuring magnitude of approximated gradient embeddings. The multi-label diversity criterion is associated to the selection of a set of images that are as diverse as possible to each other that prevents redundancy among them. To assess this criterion, we exploit a clustering based strategy. We combine each of the above-mentioned uncertainty strategies with the clustering based diversity strategy, resulting in three different query functions. All the considered query functions are introduced for the first time in the framework of MLC problems in RS. Experimental results obtained on two benchmark archives show that these query functions result in the selection of a highly informative set of samples at each iteration of the AL process.
\end{abstract}
\IEEEpeerreviewmaketitle

\section{Introduction}
%
%
%
%
\IEEEPARstart {T}{he} development of multi-label image classification (MLC) methods (which aim to automatically assign multiple land-use land-cover class labels [i.e., multi-labels] to each image scene in a remote sensing (RS) image archive) is a growing research interest in RS \cite{8089668}. In recent years, deep neural networks (DNNs) have attracted great attention for MLC of RS images due to their high capability to describe the complex spatial and spectral content of RS images. For effectively learning the model parameters of DNNs, a sufficient number of high-quality (i.e., reliable) training images annotated with multiple class labels is required. However, the collection of multi-labels is highly costly in terms of human time and effort \cite{zegeye2018novel}. To reduce annotation efforts, active learning (AL), which aims at iteratively expanding an initial labeled training set based on interaction between the user and the classification system, can be used. At each AL iteration, for a given classifier the most informative unlabeled images (i.e., samples) are selected based on a function (i.e., query function). Then, these samples are labeled by a supervisor and added to the current training set. The supervisor is usually a human expert, who annotates the selected samples with the true class labels. The iterative process is terminated once performance converges or a predefined budget limit is reached. 

Most of the existing AL methods for DNNs are developed for pixel-based classification (i.e., semantic segmentation) problems for land-cover maps generation or change detection~\mbox{\cite{9774342, 9756343,9136674}} in RS. For MLC problems, AL is seldom considered in RS~\mbox{\cite{zegeye2018novel, 9832017}}. An AL technique is introduced in~\mbox{\cite{zegeye2018novel}} for multi-label support vector machines. It exploits a query function based on the evaluation of two criteria: i) multi-label uncertainty; and ii) multi-label diversity. The multi-label uncertainty criterion is associated to the confidence of the MLC algorithm in correctly assigning multi-labels to the considered sample, while the multi-label diversity criterion aims at selecting a set of uncertain samples that are as diverse as possible, and thus reducing the redundancy among the selected uncertain samples. As that technique defines the informativeness of samples based on traditional supervised classifiers with hand-crafted features, its query function is not suitable to be directly used with DNNs. For MLC of RS images through DNNs, an AL strategy is introduced in~\mbox{\cite{9832017}} to reduce annotation errors by partially annotating images with salient labels. Instead of selecting the most informative samples to be annotated, in that strategy each sample is partially annotated with class labels, for which the confidence of the considered DNN is high. This can lead to unnecessary and redundant labeling of non-informative images, and thus an increase in labeling cost and a limited DNN model capacity.

The development of DNN-based AL methods is much more extended in the computer vision (CV) community with a main focus on single-label classification or semantic segmentation problems. As an example, in~\mbox{\cite{sener2018active}} a core set approach based on a greedy k-centre like algorithm is proposed to find a set of diverse samples. It has been found to be an effective strategy for sample diversity if the number of classes are not high. In~\mbox{\cite{gal2017deep}}, an AL framework is introduced as an ensemble method to measure the prediction confidence of DNNs through stochastic forward passes for assessing sample uncertainty. Although such ensemble methods are found to be successful for AL, they suffer from slow sample selection due to their requirement for multiple forward passes on DNNs. In recent years, informative sample selection through the estimation of DNN loss functions has attracted increasing attention for AL in CV~\mbox{\cite{Yoo_2019_CVPR,Huang_2021_ICCV}}. We refer readers to~\mbox{\cite{ren2021survey}} for a detailed review of AL methods in CV.

In this letter, we take the first step to introduce deep AL for selecting the most informative unlabeled images to be annotated with multi-labels, and thus minimizing annotation efforts of MLC problems in RS. In this context, we investigate the effectiveness of three different AL query functions designed to assess the multi-label uncertainty and multi-label diversity.

\section{Investigated Query Functions for AL in MLC}
\label{methodology}
Let $\mathcal{X}=\{\mathbf{X}_1,\ldots,\mathbf{X}_N\}$ be an archive of $N$ RS images (i.e., samples), where $\mathbf{X}_n$ is the $n$th image in the archive. We assume that an initial training set $ \mathcal{T}^{1}$ with $M$ samples annotated with multi-labels $\mathcal{Y}^{1} = \{\mathbf{y}_1,\ldots,\mathbf{y}_{M}\}$ is available. $\mathbf{y}_j = [y^{j}_1,\ldots,y^{j}_C] \in \{0,1\}^{C}$ is the multi-label vector of the image $\mathbf{X}_j \in \mathcal{T}^{1}$ that shows which classes are present in the image, where $C$ is the number of classes. AL aims to iteratively enrich the training set from a large set of unlabeled samples by selecting the most informative samples. In the context of DNNs, at every iteration $\tau$ (starting with $\tau=1$) a DNN $\mathcal{F}$ is trained using the training set $\mathcal{T}^{\tau}$, and then $b$ samples are selected from the set $\mathcal{U}^{\tau} = \mathcal{X} \setminus \mathcal{T}^{\tau}$ of unlabeled samples based on a query function that finds the most informative samples for $\mathcal{F}$. The value of $b$ is associated to labeling budget per iteration. The samples selected through the query function are labeled by a human expert and added to the training set to form the enriched training set $\mathcal{T}^{\tau+1}$ with labels $\mathcal{Y}^{\tau+1}$. 
In this paper, we investigate three AL query functions that are based on the evaluation of two criteria: i) multi-label uncertainty; and ii) multi-label diversity. These criteria are applied in two consecutive steps as follows. The most uncertain $mb$ samples are selected in the first step, and then $b$ uncertain samples (which are as diverse as possible to each other) are selected in the second step, where $m$ is a hyperparameter. For training $\mathcal{F}$, we exploit the binary cross entropy (BCE) loss function, which is defined as follows:
\begin{equation}
    \mathcal{L}_{BCE}(\mathcal{F}(\mathbf{X}_j), \mathbf{y}_j) = -\!\!\sum^C_{i=1} y^{j}_i \log( p^{j}_i)\! + \!(1\!-\!y^{j}_i) \log(1\!-\!p^{j}_i),
\end{equation}
where $[p^{j}_1,\ldots,p^{j}_C]$ is the class probability vector obtained from $\mathcal{F}(\mathbf{X}_j)$. In the subsections \ref{uncertainty} and \ref{diversity}, the strategies for evaluating multi-label uncertainty and multi-label diversity are presented.

\subsection{Evaluation of the Multi-Label Uncertainty} \label{uncertainty}
The multi-label uncertainty of an unlabeled sample is associated to the confidence of the considered DNN in correctly assigning multi-labels to the considered sample. To measure the multi-label uncertainty in the framework of DNNs, we investigate three different strategies: 1) learning multi-label loss ordering (denoted as LL); 2) measuring temporal discrepancy of multi-label prediction (denoted as TPD); and 3) measuring the magnitude of approximated gradient embeddings (denoted as MGE).
\subsubsection{The LL strategy}
It aims to measure the multi-label uncertainty through the estimation of potential losses. To this end, it exploits a subnetwork to predict the ordering of unlabeled samples based on their potential losses inspired by~\mbox{\cite{Yoo_2019_CVPR}}. Then, high-ranked samples in the ordering are assumed to produce a high loss. Such samples are considered to be uncertain and selected for annotation. In detail, in addition to $\mathcal{F}$ we employ an auxiliary subnetwork $\mathcal{F}_L$ connected to $\mathcal{F}$. For the $j$th training sample $\mathbf{X}_j$, the hidden features from $\mathcal{F}$ (which are denoted by $\mathcal{F}_{hidden}(\mathbf{X}_j)$) are fed to $\mathcal{F}_L$. In $\mathcal{F}_L$, these features are transformed using global average pooling followed by a fully connected layer with ReLU activation function that results in the prediction of the multi-label loss ranking $\hat{l}_j = \mathcal{F}_L(\mathcal{F}_{hidden}(\mathbf{X}_j))$. $\mathcal{F}$ and $\mathcal{F}_L$ are trained jointly. Let $B^p$ be the set of training sample pairs, which are randomly selected from a given mini-batch $B$. For a pair ($\mathbf{X}_j, \mathbf{X}_k) \in  B^p$, the loss function for training $\mathcal{F}_L$ is defined as follows: 
\begin{equation}
    \mathcal{L}_{\mathcal{F}_L}((l_j, l_k), (\hat{l}_j, \hat{l}_k)) \!=\! \max (0,\!-sign(l_j \!-\!l_k)(\hat{l}_j\! - \!\hat{l}_k) +\xi),
\end{equation}
where $l_i = \mathcal{L}_{BCE}(\mathcal{F}(\mathbf{X}_i),\mathbf{y}_i)$ denotes the true loss of a sample $\mathbf{X}_i$ and $\xi > 0$ is the margin hyperparameter. The loss $\mathcal{L}_{\mathcal{F}_L}$ is high if the order of the predicted values is not inline with the order of the true losses (e.g., $\hat{l}_j > \hat{l}_k$ while $l_k > l_j$). By this way, $\mathcal{F}_L$ learns the relative relations of the samples based on their multi-label loss values that can be used to rank samples. For $B$, the overall loss function for jointly training $\mathcal{F}$ and $\mathcal{F}_L$ is written as follows:
\begin{equation}
    \mathcal{L} = \frac{1}{|B|}\sum_{\mathbf{X}_i \in B} l_i + \lambda \frac{2}{|B|}\!\!\!\!\!\!\!\!\!\!\! \sum_{\quad\quad(\mathbf{X}_j,\mathbf{X}_k) \in B^p} \!\!\!\!\!\!\!\!\!\!\!\mathcal{L}_{\mathcal{F}_L}((l_j,l_k),(\hat{l}_j, \hat{l}_k)),
\end{equation}
where $\lambda$ is a hyperparameter controlling the contribution of $\mathcal{F}_L$ to the overall training. Once $\mathcal{F}$ and $\mathcal{F}_L$ are jointly trained, the uncertainty of an unlabeled sample $\mathbf{X}_j$ is evaluated as follows:
\begin{equation}
    Unc^{LL}(\mathbf{X}_j) = \hat{l}_j.
\end{equation}

\subsubsection{The TPD strategy}
It aims to measure the multi-label uncertainty of samples through temporal discrepancies of samples. To this end, the temporal discrepancy of a sample is measured based on the DNN outputs from two successive AL iterations inspired by~\mbox{\cite{Huang_2021_ICCV}}. In this strategy, if the considered DNN is already confident in predicting samples during previous AL iterations, the outputs of the DNN for such samples are assumed to be more stable during successive AL iterations. This condition is associated with small sample discrepancies and less uncertain samples. In detail, let $\mathcal{F}^{\tau}$ be the DNN model obtained after the $\tau$th iteration of the AL process and $\mathbf{X}_j \in \mathcal{U}^{\tau}$ be an unlabeled sample. Then, the multi-label uncertainty of $\mathbf{X}_j$ at the $\tau$th iteration is computed based on the temporal prediction discrepancy as follows:
\begin{equation}
  Unc^{TPD}(\mathbf{X}_j)  = \| \mathcal{F}^{\tau}(\mathbf{X}_j) - \mathcal{F}^{\tau-1}(\mathbf{X}_j) \|_2.
\end{equation}
It is noted that in the first iteration ($\tau=1$), there is no reference DNN $\mathcal{F}^{\tau-1}$ available. In this case, this strategy relies on random sampling.

\subsubsection{The MGE strategy}
It aims to estimate multi-label uncertainty by exploiting gradient approximations. Since the gradient magnitude of a loss function determines the extent of the model parameter change of a DNN, a high value of the gradient magnitude for $\mathcal{L}_{BCE}$ is regarded as the indicator of uncertainty in this strategy. To approximate the gradients with respect to the penultimate layer of the considered DNN, we utilize the pseudo-labels inspired by~\mbox{\cite{ashdeep}}. Let $\hat{\mathbf{y}}_j = [\hat{y}^j_1,\ldots,\hat{y}^j_C] \in \{0,1\}^C$ be the pseudo-label for $\mathbf{X}_j$ obtained by thresholding the output probabilities $\mathcal{F}(\mathbf{X}_j)$ with a value of $0.5$. The approximation of the gradients with respect to the penultimate layer of $\mathcal{F}$ for the loss $\mathcal{L}_{BCE}$ is defined as follows: 
\begin{equation}
g_{\mathbf{X}_j}^{\hat{\mathbf{y}}_j} = \nabla_W \mathcal{L}_{BCE}(\mathcal{F}(\mathbf{X}_j),\hat{\mathbf{y}}_j),
\end{equation}
where $W$ is the set of the parameters of the last layer in $\mathcal{F}$. {For all the unlabeled samples, we compute the gradient approximations associated with them and utilize the corresponding magnitude values to assess their uncertainty.} In detail, the multi-label uncertainty of an unlabeled sample based on the magnitude of approximated gradient embeddings is defined as follows:
\begin{equation}
    Unc^{MGE}(\mathbf{X}_j) = \|g_{\mathbf{X}_j}^{\hat{\mathbf{y}}_j}\|_2.
\end{equation}
\subsection{Evaluation of Multi-Label Diversity} \label{diversity}
To assess the diversity of unlabeled samples, we exploit a clustering-based strategy as it has been found an effective way in RS \cite{5611590} due to the fact that unlabeled uncertain images from different clusters are implicitly sparse in the feature space. To this end, the most uncertain $mb$ samples (which are selected using one of the above-mentioned strategies to measure the multi-label uncertainty) are clustered into $b$ clusters by using the Kmeans++ algorithm. Then, from each cluster we select the sample associated to the highest multi-label uncertainty that results in $b$ selected samples. Due to selection of one sample from each cluster, the multi-label diversity of samples at each AL iteration is achieved. 

We combine each multi-label uncertainty strategy with the clustering based multi-label diversity strategy. This results in three different query functions investigated in this paper: 1) LL with Clustering (denoted by LL+Clustering); 2) TPD with Clustering (denoted by TPD+Clustering); and 3) MGE with Clustering (denoted by MGE+Clustering).

\section{Experimental Results}
\subsection{Dataset Description and Design of Experiments}
Experiments were conducted on the UCMerced \cite{6257473} and TreeSatAI \cite{treesat} datasets. The UCMerced dataset consists of 2100 RGB images, each of which has the size of 256 $\times$ 256 pixels with a spatial resolution of 30cm. Each image in this dataset is annotated with multi-labels in \cite{8089668}, while the total number of classes is 17. The images (i.e., samples) were randomly divided to derive a validation set of 525 samples, a test set of 525 samples, and a pool of 1050 samples. From the pool, 40 images were randomly chosen to construct the initial training set and the rest were considered as unlabeled samples. The TreeSatAI dataset is made up of 50381 aerial images, each of which has the size of 302 $\times$ 302 pixels and include RGB and near infrared bands. The images in this dataset were resized to 288 $\times$ 288 pixels for the experiments. Each image is annotated with multi-labels, while the total number of classes is 15 and each class label is associated with an area coverage ratio. We retained a class label if its coverage ratio is higher than 0.07. This resulted in the derivation of a validation set of 6799 samples, a test set of 5043 samples, and a pool of 38531 samples. From the pool, 1000 images were randomly chosen to construct the initial training set and the rest are considered as unlabeled samples.

We report our results in terms of $F_1$ scores. All experimental results are referred to the average accuracies obtained in three trials, while an initial training set was randomly reconstructed in each trial. We used Densenet121 architecture for $\mathcal{F}$. For training, we utilized the stochastic gradient descent (SGD) optimizer with an initial learning rate of 0.025 and 0.01 on UCMerced and TreeSatAI, respectively. For UCMerced, training was conducted for 100 epochs with the mini-batch size of 10. For TreeSatAI, we increased the number of epochs to 200 and used a mini-batch size of 100. We reduced the learning rate by a factor of $0.1$ after 80 and 160 epochs on UCMerced and TreeSatAI, respectively. For the images of both datasets, we applied horizontal flipping and random rotation of $r \in \{0,90,180,270\}$ degrees. The value of the clustering parameter $m$ is set to 3 and 4 for the UCMerced and TreeSatAI datasets, respectively. For LL, $\xi$ was set to $1$ by following the suggestion in \cite{Yoo_2019_CVPR}. For the auxiliary subnetwork of LL, we used the Adam optimizer with the same initial learning rate with $\mathcal{F}$. The gradient propagation of the subnetwork to $\mathcal{F}$ was stopped after 60 and 120 epochs for UCMerced and TreeSatAI, respectively, as suggested in \cite{Yoo_2019_CVPR}. 
\begin{figure}[t]
\centering
\includegraphics[width=6.6cm]{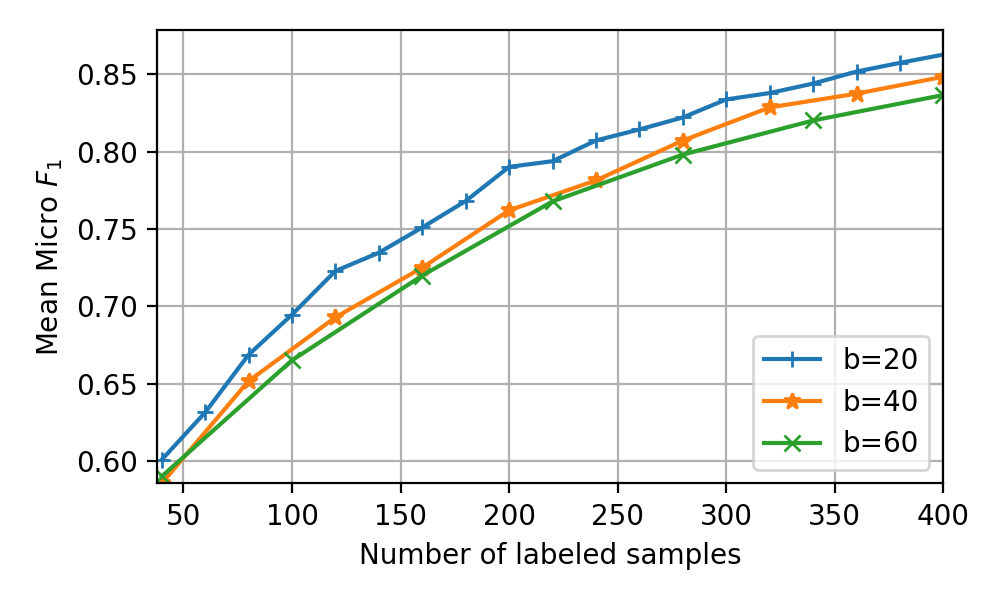}
\caption{Micro $F_1$ score versus number of training samples obtained by LL+Clustering with different $b$ values (UCMerced).}
\label{fig:adding}
\end{figure}
\subsection{Sensitivity Analysis} 
\begin{figure}[t]
\centering
\includegraphics[width=6.6cm]{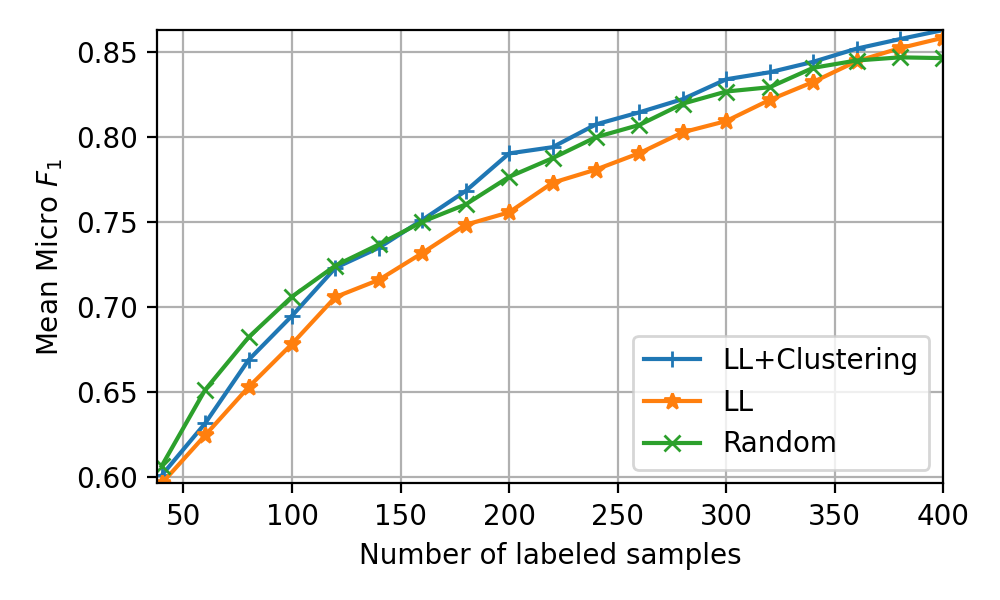}
\caption{Micro $F_1$ score versus number of training samples obtained by LL and LL+Clustering when $b$=20 (UCMerced).}
\label{fig:diversity}
\label{UCMerced_ablation}
\end{figure}

In this subsection, we initially analyze the performances under different values of the labeling budget $b$ by fixing the query function. Fig. \ref{fig:adding} shows the $F_1$ score versus number of training samples obtained by LL+Clustering query function for the UCMerced dataset. From the figure, one can observe that selecting small $b$ values results in higher $F_1$ scores under the same number of labeled samples. Furthermore, in the case of using small $b$ values higher $F_1$ scores are achieved with less samples compared to the cases of larger $b$ values. However, since at each AL iteration the DNN has to be trained on all the labeled samples, small $b$ values are associated to the higher training complexity. It is noted that similar behaviors are observed for the other query functions applied to both datasets (not provided due to space limitations). We also analyze accuracies obtained by using only multi-label uncertainty criterion and its combination with multi-label diversity criterion. Fig. \ref{fig:diversity} shows the average $F_1$ score versus number of training samples obtained by LL and LL+Clustering on the UCMerced dataset when $m$ and $b$ are set to 1 and 20, respectively. By analyzing the figure, one can observe that the classification performances are significantly improved by using both uncertainty and diversity criteria under each number of labeled samples. These results are also confirmed in the experiments with the other query functions applied to both datasets (not reported for space constraints).

\begin{figure}[t]
    \newcommand{\figwidth}{\linewidth}
    \centering
     \begin{minipage}[t]{\figwidth}
        \centering
        \centerline{\includegraphics[width=6.6cm]{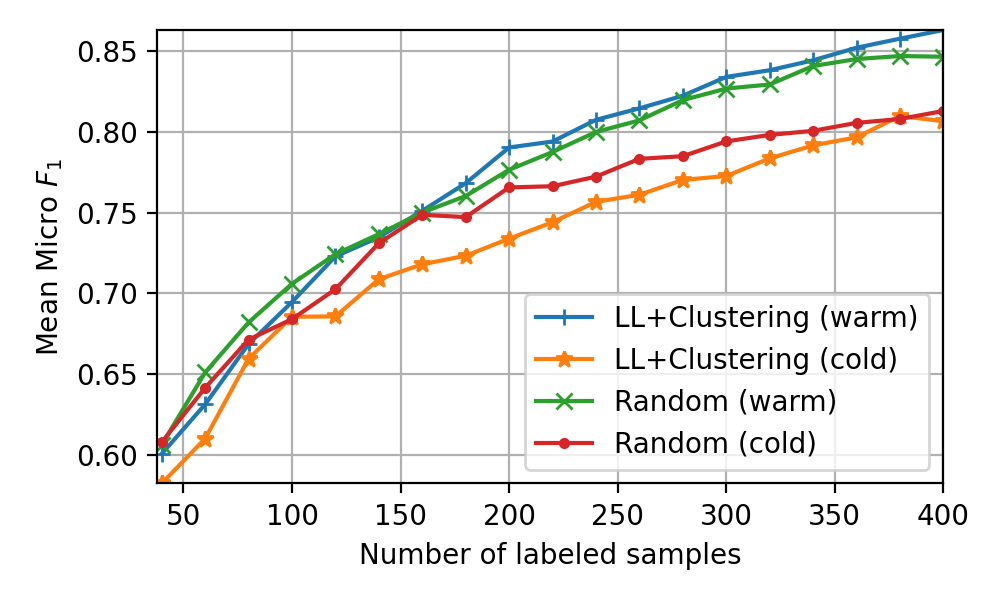}}
        \vspace{-0.3cm}
        \centerline{\footnotesize(a)}
    \end{minipage}
     \begin{minipage}[t]{\figwidth}
        \centering
        \centerline{\includegraphics[width=6.6cm]{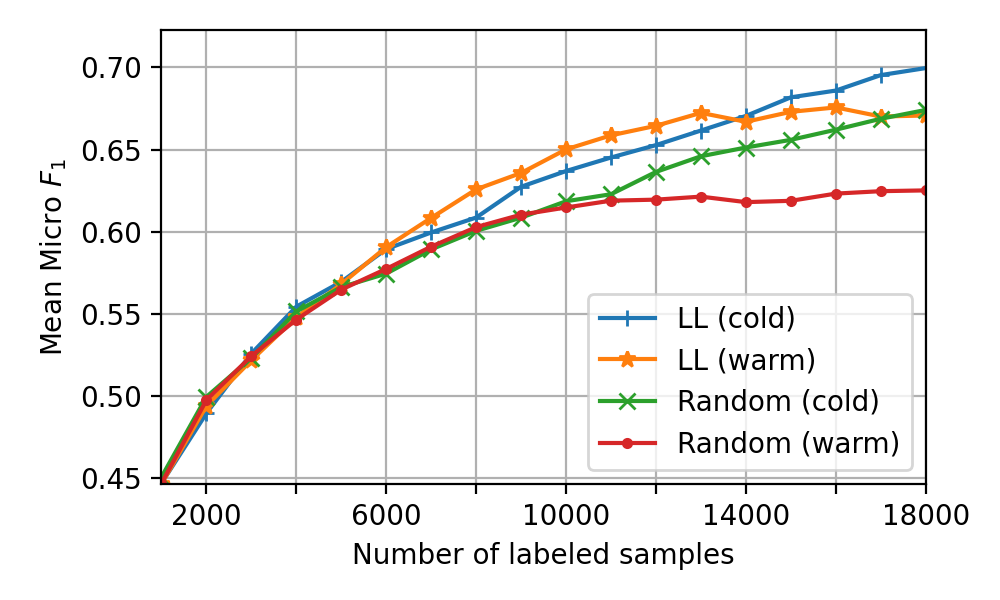}}
        \vspace{-0.3cm}
        \centerline{\footnotesize(b)}
    \end{minipage}
    \caption{Micro $F_1$ scores obtained by LL+Clustering on (a) UCMerced and (b) TreeSatAI when cold-starting and warm-starting are used.}
\label{warm_cold}
\end{figure}

The last analysis of this subsection is associated to the initialization of the model parameters. There are two different strategies for initializing the model parameters at each AL iteration: 1) random initialization that is associated to cold-starting; and 2) using the parameters learned in the previous AL iterations (except in the first iteration where the parameters are randomly initialized) that is associated to warm-starting. Fig. \ref{warm_cold} shows the mean $F_1$ scores for LL+Clustering on UCMerced and TreeSatAI using warm-starting and cold-starting. From the figure, one can see that using warm-starting results in higher scores on the UCMerced dataset (Fig. \ref{warm_cold} (a)). In the first few iterations, the model achieves already a high performance, which means that the learned weights already capture valuable knowledge about the data. Therefore, the model can benefit from re-initialization with previously learned weights. However, the figure shows that cold-starting leads to higher $F_1$ scores compared to warm-starting for TreeSatAI (Fig. \ref{warm_cold} (b)). In addition, at the first iterations warm-starting shows no benefit, whereas during subsequent iterations, when the model already achieves relatively good results, there are minor benefits. However, at later iterations warm-starting leads the model to overfit, and thus results in low accuracies. This is due to the accumulation of network updates over the course of the AL iterations.

\begin{figure}[t]
\centering
\includegraphics[width=6.6cm]{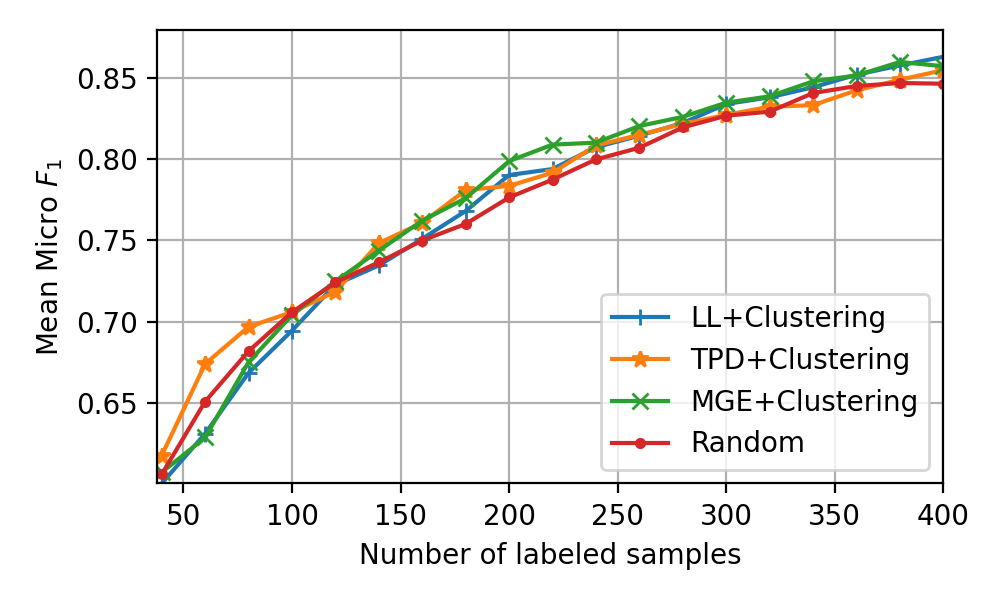}
\caption{Micro $F_1$ score versus number of labeled samples obtained by our query functions and random sampling using warm-starting (UCMerced).}
\label{UCMerced_micro}
\end{figure}
\subsection{Comparison Among the Investigated Query Functions}
\begin{figure}[t]
    \newcommand{\figwidth}{\linewidth}
    \centering
     \begin{minipage}[t]{\figwidth}
        \centering
        \centerline{\includegraphics[width=6.6cm]{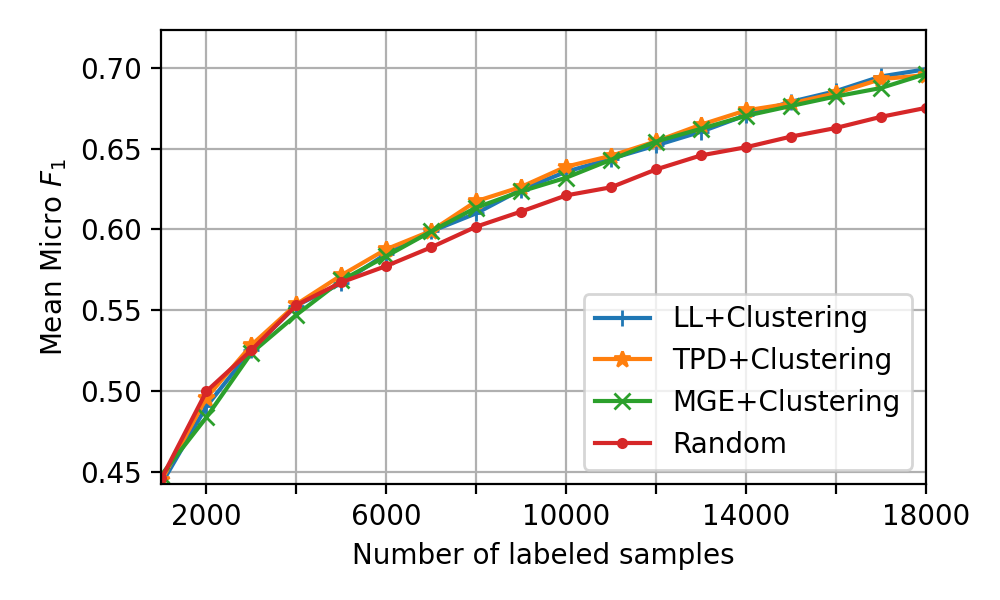}}
        \vspace{-0.3cm}
        \centerline{\footnotesize(a)}
    \end{minipage}
     \begin{minipage}[t]{\figwidth}
        \centering
        \centerline{\includegraphics[width=6.6cm]{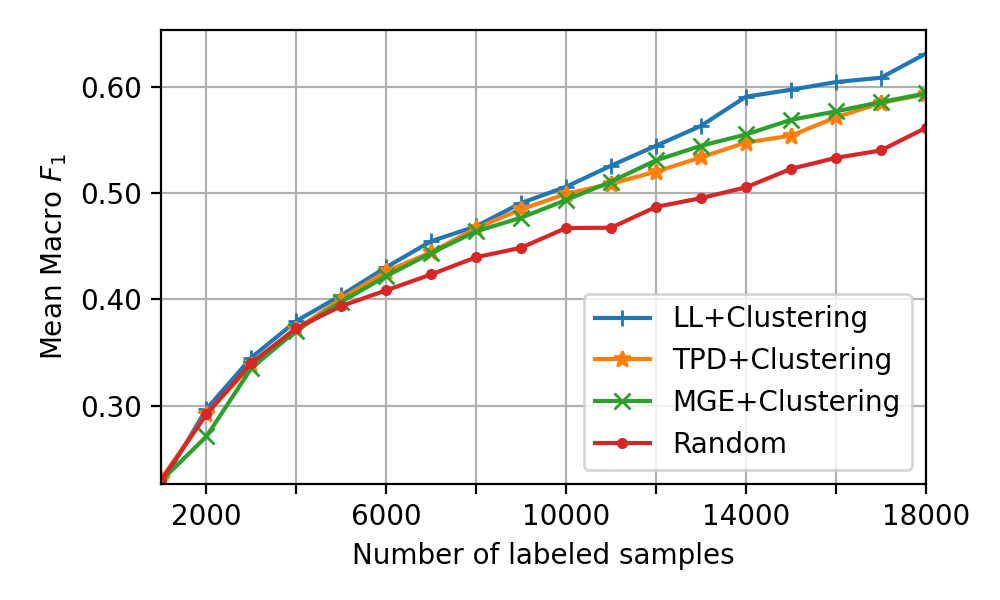}}
        \vspace{-0.3cm}
        \centerline{\footnotesize(b)}
    \end{minipage}
    \caption{(a) Micro $F_1$ and (b) macro $F_1$ scores versus number of labeled samples obtained by our query functions and random sampling using cold-starting (TreeSatAI).}
\label{TreeSat_comparison}
\end{figure}

In this section, we compare our query functions (LL+Clustering, TPD+Clustering, MGE+Clustering) among each other and with random sampling in terms of classification accuracy and complexity (associated to the trainable parameters and time needed for sampling). Fig. \ref{UCMerced_micro} shows the micro $F_1$ score versus number of labeled samples for the UCMerced dataset using warm-starting. From the figure, one can see that all our query functions provide similar performances. However, MGE+Clustering generally results in slightly higher micro $F_1$ scores. In addition, our query functions outperform random sampling, particularly at the later AL iterations. As an example, MGE+Clustering outperforms random sampling by more than 2\% reaching a micro $F_1$ score of 79.87\% with 200 labeled samples, and more than 1\% reaching a micro $F_1$ score of 85.81\% with 400 labeled samples. As another example, LL+Clustering outperforms random sampling by more than 1\% with a micro $F_1$ score of 86.29\% at 400 labeled samples. We have found similar results in terms of macro $F_1$ score, but could not report for space constraints. Fig. \ref{TreeSat_comparison} shows the $F_1$ score versus number of training samples obtained for TreeSatAI using cold-starting. From the figure, one can observe that the three query functions show similar performances, while outperforming random sampling in terms of micro $F_1$ score (Fig. \ref{TreeSat_comparison} (a)). For example, LL+Clustering outperforms random sampling by more than 2\% reaching a micro $F_1$ score of 69.90\% with 18000 labeled samples. From Fig. \ref{TreeSat_comparison} (b), one can see that LL+Clustering provides the highest macro $F_1$ scores. LL+Clustering outperforms random sampling by 8\% reaching a macro $F_1$ score of 59.05\% with 14000 labeled samples. MGE+Clustering outperforms random sampling by about 5\% with 14000 labeled samples reaching a macro $F_1$ scores of 55.49\%, while TPD+Clustering outperforms random sampling by more than 4\% reaching a macro $F_1$ score of 54.74\%. Table \ref{time_table} shows the sampling time and the number of trainable parameters for each method. LL+Clustering has the highest number of trainable parameters as well as hyperparameters, which may require some additional effort for tuning. The time for selecting the most informative samples at one AL iteration are similar for LL+Clustering and MGE+Clustering, whereas TPD+Clustering is slightly slower due to the computation overhead of the reference model.

\section{Conclusion}
In this letter, we have investigated three different AL query functions in the framework of DNNs for the MLC of RS images. The presented query functions are based on the evaluation of two criteria: i) multi-label uncertainty; and ii) multi-label diversity. To evaluate multi-label uncertainty, we have investigated the effectiveness of three different strategies: i) learning multi-label loss ordering; ii) measuring temporal discrepancy of multi-label predictions; and iii) measuring magnitude of approximated gradient embeddings. To achieve multi-label diversity, we have exploited a clustering based strategy. We have combined each multi-label uncertainty evaluation strategy with the clustering based strategy, resulting in three different query functions. It is worth emphasizing that our query functions are independent from the considered DNN and can be used with any DNN architecture designed for MLC problems. From the experimental results, we have observed that the combination of learning multi-label loss ordering with clustering leads to the best overall performance, whereas that of measuring magnitude of approximated gradient embeddings is computationally more efficient with a slight decrease in the accuracy. As a future work, we plan to adapt and test our query functions in the framework of multi-label RS image retrieval problems.
\begin{table}
\renewcommand{\arraystretch}{1.2}
    \centering
    \caption{Time Needed for Selecting $b$ Unlabeled Samples and the Number of Trainable Parameters for Each AL Query Function When $b$=20 for UCMerced and $b$=1000 for TreeSatAI}
    \begin{tabular}{@{}lccc@{}}
    \cline{1-4} 
        \multirow{2}{0.11\textwidth}[0cm]{Query Function} & \multicolumn{2}{c}{Time} & \multirow{2}{0.07\textwidth}[0cm]{Number of Parameters} \\ \cline{2-3}
        & UCMerced & TreeSatAI & \\   
        \cline{1-4} 
        LL + Clustering & 7.25s & 141.5s & 7.33M\\
        TPD + Clustering & 10.84s & 188.1s & 6.97M\\
        MGE + Clustering & 7.06s& 142.2s & 6.97M\\
        \cline{1-4} 
    \end{tabular}
    \label{time_table}
\end{table}

\bibliographystyle{IEEEtran}
\bibliography{references.bib}
\end{document}